# The corruptive force of AI-generated advice

Margarita Leib[1][*][†], Nils C. Köbis[2][*][†], Rainer Michael Rilke[3], Marloes Hagens[1], Bernd Irlenbusch[*][4]

**Affiliations:**

[1] University of Amsterdam, Center for Research in Experimental Economics and Political Decision Making (CREED), University of Amsterdam, Roetersstraat 11, Postbus 15867, 1001 NJ Amsterdam, The Netherlands [2] Max Planck Institute for Human Development, Center for Humans and Machines, Lentzallee 94, 14195 Berlin. [3] WHU - Otto Beisheim School of Management, Burgplatz 2, 56179 Vallendar, [4] University of Cologne.

*Corresponding authors: koebis@mpib-berlin.mpg.de, m.leib@uva.nl, bernd.irlenbusch@uni-koeln.de.

[†] These authors contributed equally to this work

CORRUPTIVE FORCE OF AI 2**Abstract:**

Artificial Intelligence (AI) is increasingly becoming a trusted advisor in people's lives. A new concern arises if AI persuades people to break ethical rules for profit. Employing a large-scale behavioural experiment ($N = 1,572$), we test whether AI-generated advice can corrupt people. We further test whether transparency about AI presence, a commonly proposed policy, mitigates potential harm of AI-generated advice. Using the Natural Language Processing algorithm, GPT-2, we generated honesty-promoting and dishonesty-promoting advice. Participants read one type of advice before engaging in a task in which they could lie for profit. Testing human *behaviour* in interaction with actual AI outputs, we provide first behavioural insights into the role of AI as an advisor. Results reveal that AI-generated advice corrupts people, even when they know the source of the advice. In fact, AI's corrupting force is as strong as humans'.

# The corruptive force of AI-generated advice

Artificial Intelligence (AI) shapes people's life on a daily basis [1]. It sets prices in online markets [2], predicts crucial outcomes, like health care needs [3], and makes recommendations ranging from entertainment content to romantic partners [4]. Increasingly, AI can become a trusted advisor, thereby affecting people's behaviour [5]. As a case in point, Amazon's chief scientist, Rohit Prasad, envisions that Alexa's role for its over 100 million users "keeps growing from more of an assistant to advisor" [6]. A new concern arises if AI persuades people to break ethical rules.

Advances in Natural Language Processing (NLP), one prominent branch of AI, render this concern more relevant than ever [7]. Trained on big data, NLP algorithms can parse text to generate advice. For example, using NLP algorithms, large companies already analyse employees' recorded sales calls and advise them how to increase their sales (e.g., gong.io). With the unconstrained objective function to maximise profits, such algorithms could recommend employees to break ethical rules. If a machine-learning algorithm, set to maximise profits, autonomously detects that deceiving customers pays off, it can advise this strategy to salespeople. Preventing such undesirable outcomes would require the designers of algorithms to be aware of its corrupting force, and explicitly constrain the algorithm from providing such harmful advice [8]. Therefore, we examine: Can AI-generated advice successfully corrupt people? That is, will people sacrifice their honesty to follow AI-generated advice?

AI-generated text increasingly appears human-like [9,10]. People might mistake an AI-generated advice as human-written and thus, unbeknownst to them, be influenced by AI. To prevent such misattribution and alleviate potential harm, governments, policy makers, and researchers recommend algorithmic transparency [11] — the mandatory disclosure of AI [12]. While being a popular policy recommendation, empirical evidence for its effectiveness to mitigate corrupting effects of AI is lacking.

Aware that the advice stems from an AI, the way in which people react to corrupting advice is not trivial. On the one hand, people voice a general aversion towards relying on algorithms [13], especially in ethical domains [14]. On top of that, people tend to follow social norms in ethically challenging situations [15–17] and advice generated by an AI is likely to be a weaker signal of a social norm compared to advice from a human. Thus, people might not follow corrupting AI-generated advice. On the other hand, when people are tempted to break ethical rules for profit, they do so as long as they can justify their actions [18,19]. Further, people deflect blame for negative outcomes not only on other people [20], but also on algorithms [21]. As such, corrupting AI-generated advice may suffice as a justification, convincing people to break ethical rules. Here, we test how informing people about the algorithmic source of advice affects their reaction to it.

**Methods**

We conducted a large-scale, financially incentivised, pre-registered experiment (https://osf.io/kns7u). In it, we employ the state-of-the-art NLP algorithm, Generative Pre-Training 2 (GPT-2) [22], to produce AI-advice texts. To facilitate reproducibility [23], the training data and code are openly available (https://github.com/marloeshagens/GPT-2). We then assess participants' (dis)honest behaviour in response to such advice. Thus, we extend prior work that has either examined stated preferences towards algorithms in hypothetical scenarios or confronted participants with human-written texts that were labelled as AI-generated. The current experiment is among the first to adopt a machine behaviour approach [1], assessing how *actual* algorithmic output influences human *behaviour*.

The design entailed two parts (Fig. 1). First, in the *advice giving task* we recruited participants (*N* = 395) to write advice. Participants were either incentivised to promote honesty or dishonesty in their advice. Using the two separate samples of human-written advice, we trained GPT-2 to generate corresponding AI-advice (see Fig. 1 Part A).

Second, in an *advice taking task,* we recruited a separate sample of participants (*N* = 1,572), who first read the instructions, then received an advice, and finally engaged in a task assessing (dis)honest behaviour. Specifically, we employed the well-established dyadic die rolling task [24], in which participants are paired in dyads, composed of a first and second mover. The first mover rolls a die in private and reports the outcome. The second mover learns about the first mover's report, rolls a die in private, and reports the outcome as well. Only if the first and second mover report the same outcome (a double), they are paid according to the double's worth, with higher doubles corresponding to higher pay. If they report different outcomes, they are not paid.

Such behavioural cheating tasks have external validity across various domains [25,26]. In a dyadic version, two moral motivations clash: telling the truth versus collaborating [24], rendering advice particularly valuable. In our novel set-up, first movers were the ones who received advice (see Fig. 1, Part B), hence, the behaviour of first movers is reported in the manuscript and second movers' behaviour is reported in the Supplementary Materials.

Before reporting the die roll outcome, participants randomly assigned to different treatments read an honesty-promoting or dishonesty-promoting advice that was either human-written or AI-generated. Further, participants either knew the source of the advice (transparency) or knew that there was a 50-50 chance that it came from either source (opacity). Participants in the opacity treatment further engaged in an incentivised version of a Turing Test [10], in which they could earn additional pay if they guessed the source of advice correctly. Lastly, as a baseline, we implemented a control treatment in which participants did not receive any advice.

The analyses in the main manuscript report linear regressions with two-tailed tests. Supplemental material on the Open Science Framework (see https://osf.io/g3sw2/) reports robustness tests, Bayesian analyses and analyses using alternative outcome variables as well as the exact *n*s for each treatment cell.

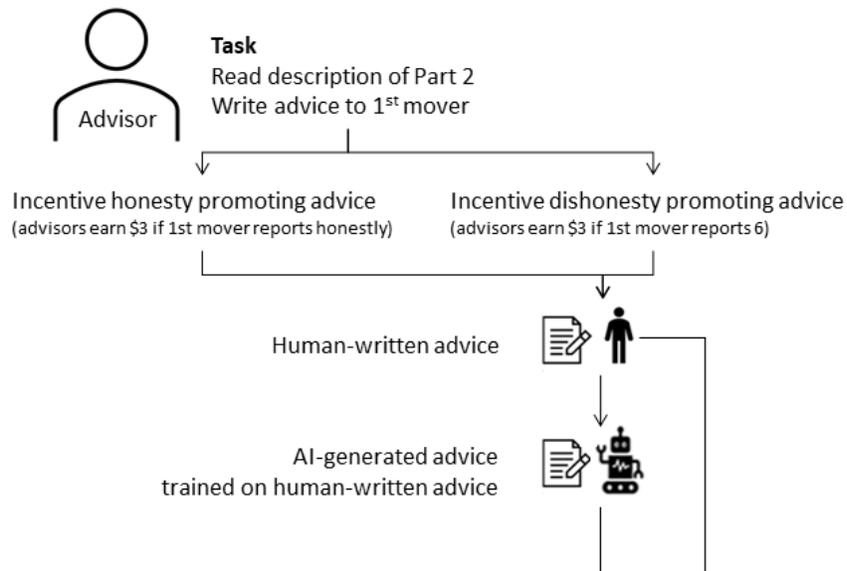

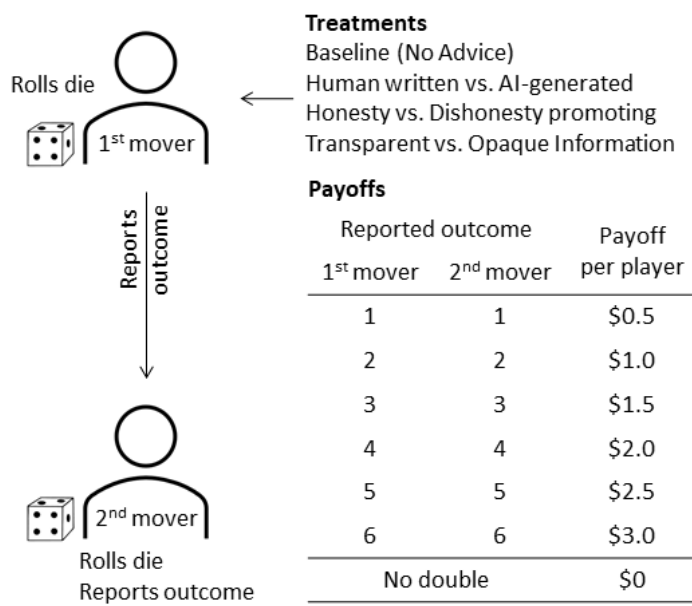

**Fig. 1**. Advice giving (A) and taking (B) tasks. (A) Participants were incentivised to write honesty or dishonesty-promoting advice. (B) Another group of participants engaged in the dyadic die rolling task. All first movers read an advice, and then reported a die roll outcome to the second mover. First movers read an honesty/dishonesty-promoting advice that was human-written or AI-generated. Participants were either informed about the source of advice (transparency) or not (opacity). As a benchmark, another group of participants did not read any advice.

**Results**

Overall, participants lied: Across all nine treatments, the average die roll outcomes exceeded the expected average if participants were honest (EV=3.5, $t$s(df>159) > 5.01, $p$s < .001). Further, when the source of the advice was not disclosed, AI-advice corrupts people. Linear regression analyses reveal that average die roll outcomes following AI-generated advice that promoted dishonesty ($M = 4.99$, $SD = 1.42$) significantly exceeded reports that followed either AI-generated advice promoting honesty ($M = 4.22$, $SD = 1.51$, $b = -.76$, $p < .001$, $d_{Cohen} = -0.53$), or no advice ($M = 4.22$, $SD = 1.56$, $b = -.77$, $p < .001$, $d_{Cohen} = -0.52$). Honesty-promoting AI-advice failed to sway people's behaviour. Namely, there was no difference in reported die roll outcomes between the AI-generated honesty-promoting advice and the no advice treatment ($p = .965$). Bayesian analyses corroborating these findings are reported in the Supplementary Materials).

Put in perspective, relative to no advice, honesty-promoting AI-advice did not change average die roll reports at all [(4.22-4.22)/4.22 = .00], and dishonesty-promoting AI-advice increased reports by 18% [(4.99-4.22)/4.22 = .18]. Moreover, when not knowing the source of the advice, the effect of AI-generated advice was indistinguishable from that of human-written advice, as indicated by lack of advice type (honesty/dishonesty-promoting) by source (AI/human) interaction, $p = .508$.

In the opacity treatments, participants' responses in the incentivised version of a Turing Test indicated that they failed to reliably distinguish AI from human advice. They correctly identified the source in 52.95% of the cases, not exceeding chance levels (50%), binomial test, $p = .128$.

In light of this inability to detect AI, would transparency alleviate the corruptive force of AI-advice? Our results suggest no. A regression analysis revealed an insignificant three-way interaction between advice type, source, and information, $p = .599$. Both, among participants who received AI-advice as well as human advice, the two-way advice type (honesty/dishonesty-promoting) by information (opacity/transparency) interaction was not significant ($p = .357$ and $p = .859$, respectively), Fig. 2.

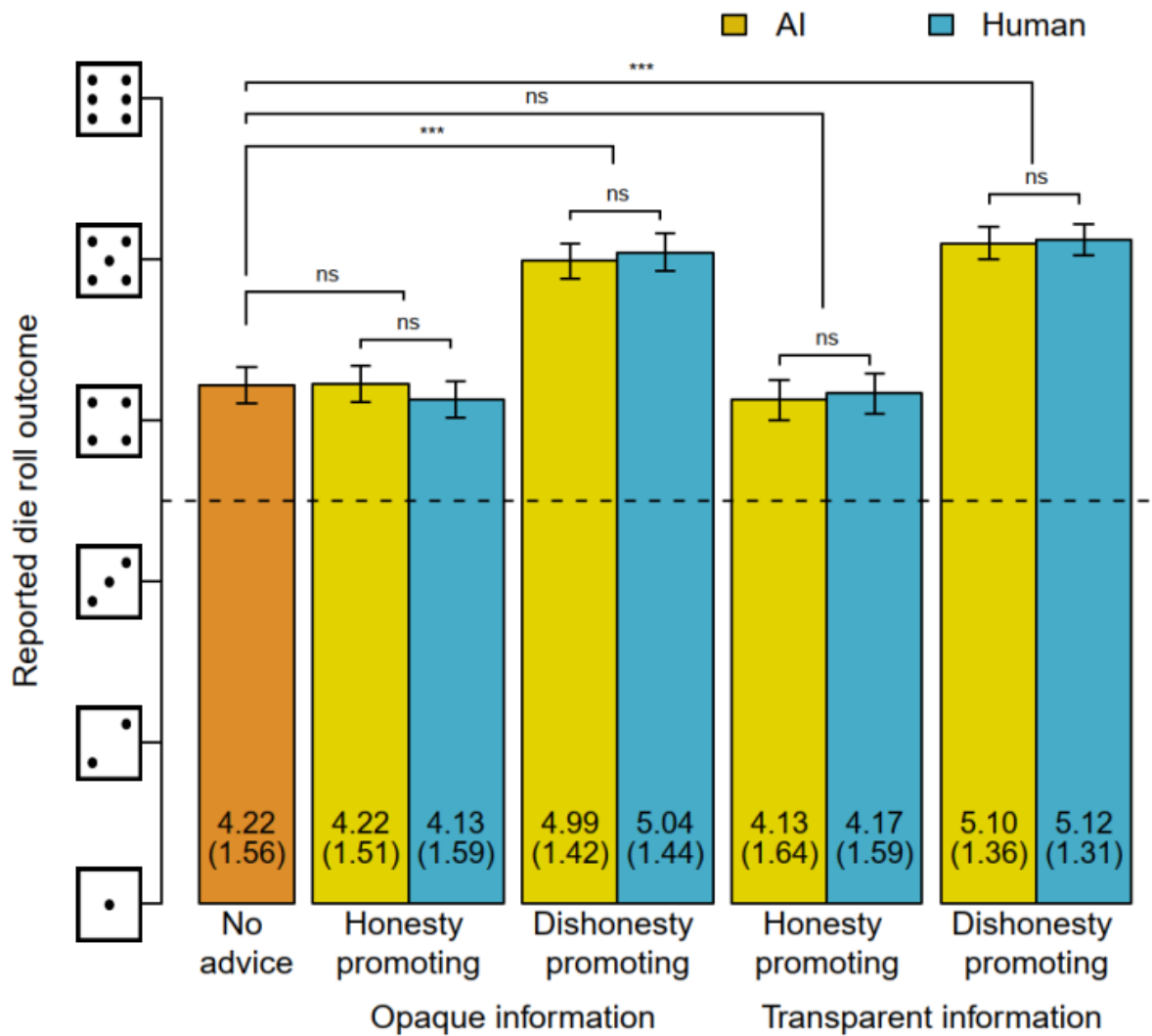

**Fig. 2.** Mean reported die roll outcomes across advice types, source, and information. Dashed line represents the expected mean if participants were honest (3.5). Mean (SD) are at the bottom of each bar, error bars indicate ±1 *SE*. \*\*\**p* < .001 ns: *p* >.05.

Since intelligent algorithms take an ever-growing role in human lives, research needs to examine their potential benefits and harms. Here we provide insights into how AI potentially contributes to (un)ethical behaviour [2]. As adherence to (ethical) rules has immense impact on the welfare of society [15], AI could be a force for good, if it manages to convince people to act more ethically. Yet our results reveal that AI advice fails to increase honesty. Instead, AI can be a force for evil: advice generated by GPT-2 drastically increased dishonesty, even when people knew that the advice stemmed from AI.

In fact, we provide first behavioural insights that in the role of an advisor, AI can corrupt people. Past research had shown that people deflect responsibility for their behaviour to advisors [27]. Our findings suggest that also AI-advisors can serve as scapegoats to which one can deflect (some of the) moral blame of dishonesty. Moreover, our results reveal that in the context of advice taking, transparency about algorithmic presence does not suffice to alleviate its potential harm, in contrast to the commonly proposed policy [11]. Even when knowing that an algorithm, not a human, crafted the advice, people followed it. The power of self-serving justifications to lie for profits seems to trump aversion towards algorithms.

Whereas previous work has revealed people's *stated* aversion towards AI making ethical decisions [14] and reluctance to follow AI advice, [28], our *behavioural* results paint a different picture. When AI-generated advice aligns with individuals' preferences to lie for profit, they gladly follow it, even when they know the source of the advice is an AI. It appears there is a discrepancy between stated preferences and actual behaviour, highlighting the necessity to study human *behaviour* in interaction with *actual* algorithmic outputs [1].

Those with malicious intentions could use the forces of AI to corrupt others, instead of doing so themselves. Whereas having humans as intermediaries already reduces the moral costs of unethical behaviour [29], using AI advisors as intermediaries is conceivably even more attractive. Compared to human advisors, AI advisors are cheaper, faster, and more easily scalable. Employing AI advisors as a corrupting force is further attractive as AI does not suffer from internal moral costs that may prevent it from providing corrupting advice to decision-makers. Furthermore, personalization of text can help to tailor the content, format and timing of the advice. As personalization is already successfully used in advertising [30], our findings likely reflect a lower bound estimate of the ability of AI to corrupt others.

Because employing AI as a corrupting agent is attractive, and since AI rivals human abilities, it is important to experimentally test the corruptive force of AI as a key step towards managing AI responsibly.

**References**

1. Rahwan, I. *et al.* Machine behaviour. *Nature* **568**, 477–486 (2019).

2. Calvano, E., Calzolari, G., Denicolò, V., Harrington, J. E., Jr & Pastorello, S. Protecting consumers from collusive prices due to AI. *Science* **370**, 1040–1042 (2020).

3. Obermeyer, Z., Powers, B., Vogeli, C. & Mullainathan, S. Dissecting racial bias in an algorithm used to manage the health of populations. *Science* **366**, 447–453 (2019).

4. Yeomans, M., Shah, A., Mullainathan, S. & Kleinberg, J. Making sense of recommendations. *Journal of Behavioral Decision Making* **32**, 403–414 (2019).

5. Grgić-Hlača, Nina Engel, Christoph Gummadi, Krishna P. Human decision making with machine advice: An experiment on bailing and jailing. *Proceedings of the ACM on Human-Computer Interaction* **3**, 1–25 (2019).

6. Strong, J. *AI Reads Human Emotions. Should it?* (MIT Technology Review, 2020).

7. Hirschberg, J. & Manning, C. D. Advances in natural language processing. *Science* **349**, 261–266 (2015).

8. Thomas, P. S. *et al.* Preventing undesirable behavior of intelligent machines. *Science* **366**, 999–1004 (2019).

9. Kreps, S., Miles McCain, R. & Brundage, M. All the News That's Fit to Fabricate: AI-Generated Text as a Tool of Media Misinformation. *Journal of Experimental Political Science* 1–14.

10. Köbis, N. & Mossink, L. D. Artificial intelligence versus Maya Angelou: Experimental evidence that people cannot differentiate AI-generated from human-written poetry. *Comput. Human Behav.* **114**, 106553 (2021).

11. Jobin, A., Ienca, M. & Vayena, E. The global landscape of AI ethics guidelines. *Nature*

CORRUPTIVE FORCE OF AI 14*Machine Intelligence* **1**, 389–399 (2019).

12. Diakopoulos, N. Accountability in algorithmic decision making. *Commun. ACM* **59**, 56–62 (2016).

13. Dietvorst, B. J., Simmons, J. P. & Massey, C. Overcoming Algorithm Aversion: People Will Use Imperfect Algorithms If They Can (Even Slightly) Modify Them. *Manage. Sci.* **64**, 1155–1170 (2018).

14. Bigman, Y. E. & Gray, K. People are averse to machines making moral decisions. *Cognition* **181**, 21–34 (2018).

15. Gächter, S. & Schulz, J. F. Intrinsic honesty and the prevalence of rule violations across societies. *Nature* **531**, 496–499 (2016).

16. Bowles, S. *The Moral Economy: Why Good Incentives Are No Substitute for Good Citizens*. (Yale University Press, 2016).

17. Fehr, E. Behavioral Foundations of Corporate Culture. *UBS International Center of Economics in Society* (2018) doi:10.2139/ssrn.3283728.

18. Fischbacher, U. & Föllmi-Heusi, F. Lies in Disguise—An Experimental Study on Cheating. *J. Eur. Econ. Assoc.* **11**, 525–547 (2013).

19. Barkan, R., Ayal, S. & Ariely, D. Ethical dissonance, justifications, and moral behavior. *Current Opinion in Psychology* **6**, 157–161 (2015).

20. Bazerman, M. H. & Gino, F. Behavioral Ethics: Toward a Deeper Understanding of Moral Judgment and Dishonesty. *Annual Review of Law and Social Science* **8**, 85–104 (2012).

21. Hohenstein, J. & Jung, M. AI as a moral crumple zone: The effects of AI-mediated communication on attribution and trust. *Comput. Human Behav.* **106**, 106190 (2020).

22. Radford, A. *et al.* Language models are unsupervised multitask learners. *OpenAI Blog* **1**, 9

(2019).

23. Hutson, M. Artificial intelligence faces reproducibility crisis. *Science* **359**, 725–726 (2018).

24. Weisel, O. & Shalvi, S. The collaborative roots of corruption. *Proc. Natl. Acad. Sci. U. S. A.* **112**, 10651–10656 (2015).

25. Dai, Z., Galeotti, F. & Villeval, M. C. Cheating in the Lab Predicts Fraud in the Field: An Experiment in Public Transportation. *Manage. Sci.* **64**, 1081–1100 (2018).

26. Cohn, A. & Maréchal, M. A. Laboratory Measure of Cheating Predicts School Misconduct. *Econ J* **128**, 2743–2754 (2018).

27. Bonaccio, S. & Dalal, R. S. Advice taking and decision-making: An integrative literature review, and implications for the organizational sciences. *Organ. Behav. Hum. Decis. Process.* **101**, 127–151 (2006).

28. Castelo, N., Bos, M. W. & Lehmann, D. R. Task-Dependent Algorithm Aversion. *J. Mark. Res.* **56**, 809–825 (2019).

29. Drugov, M., Hamman, J. & Serra, D. Intermediaries in corruption: an experiment. *Exp. Econ.* **17**, 78–99 (2014).

30. Matz, S. C., Kosinski, M., Nave, G. & Stillwell, D. J. Psychological targeting as an effective approach to digital mass persuasion. *Proc. Natl. Acad. Sci. U. S. A.* **114**, 12714–12719 (2017).

**Acknowledgments** We thank Yulia Litvinova, Ann-Kathrin Blanke, and Toan Huynh for research assistance.

**Funding:** The research was funded by Germany's Excellence Strategy EXC 2126/1-390838866: 'ECONtribute: Markets and Public Policy', the European Research Council (ERC-StG-637915) and the Chamber of Commerce and Industry (IHK) Koblenz.

**Author contributions:** NK, ML & RR initiated the study. All authors contributed to developing the initial idea further and designing the experiment. MH programmed and conducted the experiment under supervision of NK and ML. MH and ML analysed the data. All authors interpreted the results. NK and ML wrote the paper with feedback from the other authors.

**Competing interests:** Authors declare no competing interests.

**Data and materials availability:** All data and materials are available on the Open Science Framework: https://osf.io/g3sw2/ . Training data and code to reproduce the text outcome generated by GPT-2 are openly available (https://github.com/marloeshagens/GPT-2).